# ChromaTag: A Colored Marker and Fast Detection Algorithm

Joseph DeGol        Timothy Bretl        Derek Hoiem

University of Illinois Urbana-Champaign

{degol2, tbretl, dhoiem}@illinois.edu

## Abstract

*Current fiducial marker detection algorithms rely on marker IDs for false positive rejection. Time is wasted on potential detections that will eventually be rejected as false positives. We introduce ChromaTag, a fiducial marker and detection algorithm designed to use opponent colors to limit and quickly reject initial false detections and grayscale for precise localization. Through experiments, we show that ChromaTag is significantly faster than current fiducial markers while achieving similar or better detection accuracy. We also show how tag size and viewing direction effect detection accuracy. Our contribution is significant because fiducial markers are often used in real-time applications (e.g. marker assisted robot navigation) where heavy computation is required by other parts of the system.*

## 1. Introduction

In this paper, we introduce ChromaTag[1], a new colored fiducial marker and detection algorithm that is significantly faster than current fiducial marker systems. Fiducial markers are artificial objects (typically paired with a detection algorithm) designed to be easily detected in an image from a variety of perspectives. They are widely used for augmented reality and robotics applications because they enable localization and landmark detection in featureless environments. However, because fiducial markers often complement large real-time systems (e.g. Camera Tracking [2, 9], SLAM [35] and Structure from Motion [8]), it is important that they run much faster than 30 frames per second. Figure 1 shows the run time of several state of the art markers, of which only ChromaTag achieves processing times significantly faster than 30 frames per second (all processing uses a 3.5 GHz Intel i7 Ivy Bridge processor).

ChromaTag achieves an orders-of-magnitude speedup with good detection performance through careful design of the tag and detection algorithm. Previous marker designs (Figure 2) typically use highly contrasting (black to white) borders [4, 7, 8, 10, 15, 24] for initial detection, but black-white edges are common in images and result in many initial false detections. IDs are decoded from the tags to verify detections, but decoding is the last step in the pipeline, so most of the time is spent rejecting false tags. Tags with distinctive color patterns can be used to limit initial false detections, but color consistency and the reduced spatial resolution of color channels (Bayer grid) create challenges for ID encoding and tag localization.

[1] http://degol2.web.engr.illinois.edu/pages/ChromaTag_ICCV17.html

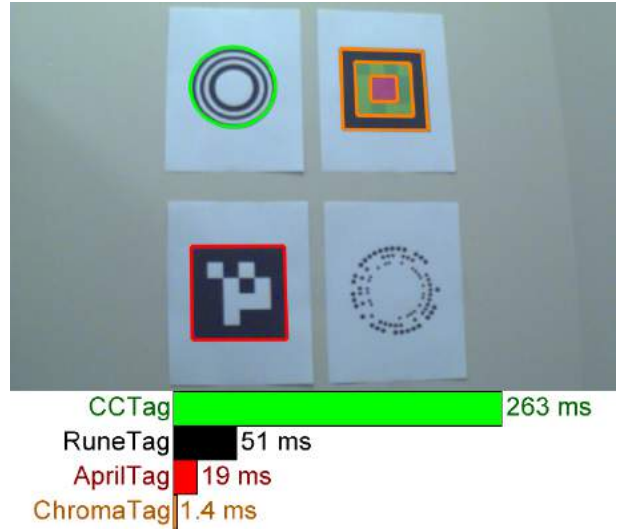

Figure 1: ChromaTag is a colored fiducial marker and detection algorithm that is significantly faster than other markers. Shown are successful detections of CCTag [8], April-Tag [33], and ChromaTag (RuneTag [4] was unsuccessful), with the time required by each tag's detection algorithm. The images in this paper are best viewed in color.

ChromaTag uses each channel of the LAB opponent colorspace to best effect. Large gradients between red and green in the A channel, which are rare in natural scenes, are used for initial detection. This results in few initial false detections that can be quickly rejected. The black-white border takes advantage of high resolution of the L channel for precise localization. The B channel is used to encode the tag ID. Our ChromaTag detection algorithm finds initial detections, builds a polygon on the borders, simplifies to a quadrilateral, and decodes the ID. Robustness to variations in lighting is achieved by using differences of chrominance and luminance throughout detection and localization. The algorithm is fast and robust and achieves precise tag localizations at more than 700 frames per second.

We collect thousands of images with ChromaTag and state-of-the-art tag designs in a motion capture arena. We use this data to demonstrate that our tag achieves significantly faster detection rates while maintaining similar or better detection accuracy for varying camera perspectives and lighting conditions. We tabulate which steps in the ChromaTag detection algorithm most often fail and consume the most time. We also evaluate how image tag size and camera viewing direction affect detection accuracy. Note that an unlabeled training video was used during algorithm design and parameter setting; the parameters are not tuned for the held out test sets.

In summary, the **contributions** of this paper are: (1) we present ChromaTag, a colored fiducial marker and detection algorithm; (2) we demonstrate that our tag achieves accurate detections faster than current state-of-the-art markers using thousands of ground truth labeled image frames with different lighting and varying perspectives; and (3) we show how ChromaTag and other fiducial markers perform as a function of image tag size and viewing direction.

## 2. Related Work

Figure 2 shows many of the tags discussed in this section. Among the earliest fiducial markers is the concentric contrasting circle (CCC) of Gatrell et al. [19] which consists of a white inner circle surrounded by a black ring with an outer white border. CCC has no signature to differentiate markers. Cho et al. [10] adds additional rings to improve detection at different depths and color to each ring as a signature. CyberCode (Rekimoto et al. [28]) is a square tag with a grid of square white and black blocks to encode signatures. Naimark et al. [23] combines the ring design of CCC with the block codes of CyberCode to create a circular marker with inner block codes. ARToolKit [1] and Fiala's [15, 16] ARTag are the first tags with widespread use for augmented reality [9]. Both tags borrow from past success with square designs by using a white square border around a black inner region. ARToolKit uses different symbols to differentiate markers while ARTag uses a grid of white and black squares. ARTag also uses hamming distance to improve false positive rejection.

Fourier Tag of Sattar et al. [30] and Xu et al. [34] is a circular tag with a frequency image as the signature which results in graceful data degradation with distance. Schweiger et al. [31] uses the underlying filters of SIFT and SURF as the design motivation for their markers, which look like

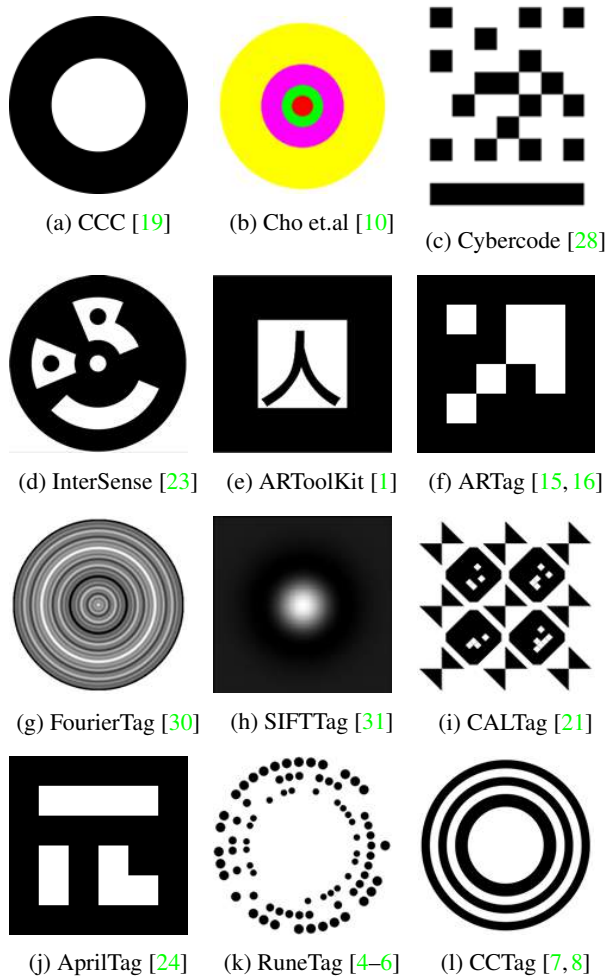

Figure 2: Existing Fiducial Marker Designs

Laplacian of Gaussian images and are specifically designed to trigger a large response with SIFT and SURF detectors. Checkerboard-based markers increase the number of corners for improved camera pose estimation: CALTag by Atcheson et al. [21] is a checkerboard with inner square markers for camera calibration and Neto et al. [13] adds color to increase the size of the signature library and remove the perspective ambiguity of checkerboard detection.

Olson and Wang's AprilTag [24, 33] is a faster and more robust reimplementation of ARTag. Garrido-Jurado et al [17, 18] uses mixed integer programming to generate additional marker codes for the square design of ARTag and AprilTag and provide their codes with the ArUco fiducial marker library [2]. Another state-of-the-art tag (RuneTag) comes from the work of Bergamasco et.al [4–6] which uses rings of dots to improve robustness to occlusion and provide more points for camera pose estimation. Lastly, CCTag by Calvet et al. [7, 8] uses a set of rings like that of Prasad et al. [26] to increase robustness to blur and ring width to encode marker signature. These tags (AprilTag, RuneTag, and CCTag) represent the current state-of-the-art for fiducial markers. They have demonstrated accurate detection and are commonly used. ChromaTag uses a new detection approach (taking advantage of color on the markers) to achieve processing times significantly faster than other tags while still maintaining similar detection accuracies.

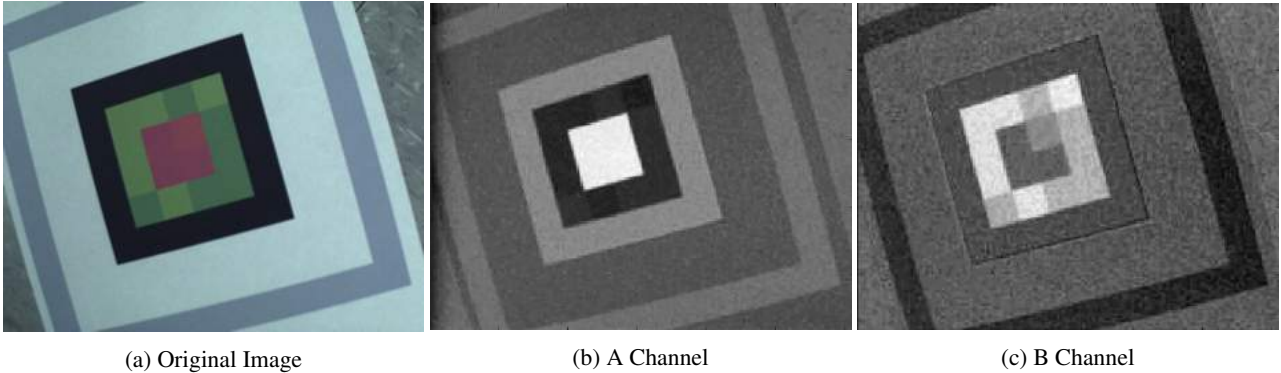

(a) Original Image        (b) A Channel        (c) B Channel

Figure 3: The original image is converted to the LAB color space and the A and B channels are shown. In the A channel, the different shades of red look the same and make a bright square region; and the different shades of green look the same and make a dark ring. This makes it easy to detect the red to green border in the A channel. In the B channel, the different shades of red and green become differentiable, making it possible to read the binary code for the tag.

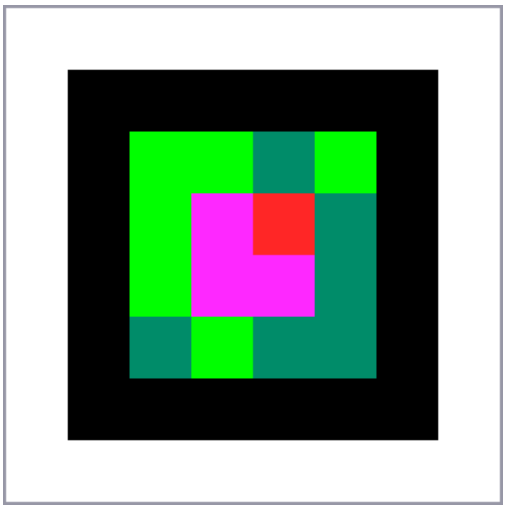

Figure 4: An example ChromaTag. The red and green regions are easily detectable and limit false positives. The two shades of red and two shades of green are used to embed the tag code. The outer black-white border provides full spatial resolution for accurate localization of the tag corners.

Several works perform planar object detection using machine learning approaches. Early work by Claus et. al [11, 12] employs a cascaded bayes and nearest neighbor classification scheme for marker detection. More recent work of Ozuysal et al. [25] and Lepetit et al. [22] uses randomized forests to learn and detect planar objects. In practice, these algorithms do not achieve detection accuracies on par with detection algorithms specifically designed for marker detection. With the increased popularity of deep learning based detection approaches, it is possible that better detection can be achieved; however, ChromaTag, which does not require a GPU, is faster than current deep learning approaches which do require a GPU [27, 29].

## 3. ChromaTag Design

Figure 4 provides an example of a ChromaTag. The inner red square and green ring are used for rejecting false positives (Section 3.1) and the outer black and white rings are used for precise localization of the tag (Section 3.2).

### 3.1. Efficient False Positive Rejection

Opponent color spaces offer large gradients between opposing colors in each channel. Figure 5 depicts the LAB color space, where red and green are opposing colors in the A channel and blue and yellow are opposing colors in the B channel. Figure 5 also depicts how two (or more) shades of green or red can have the same value in the A channel, but different values in the B channel. ChromaTag's color configuration is designed based on these properties.

The red center surrounded by the green ring has a large gradient in the A channel (Figure 3b), which rarely occurs in natural scenes (empirically validated in Figure 10). Thus, we can quickly detect tags with high precision and recall by scanning the image in steps of $N$ pixels and thresholding the A channel difference of neighboring steps.

ChromaTag encodes the binary code in the B channel (Figure 3c), which has little effect on the A channel intensity (Figure 3b). Since every tag includes low and high B values (encoding 0 and 1) in both the green and red area, the thresholds are adapted per tag to account for variations due to lighting or printing. The tag detection is verified based on the code value, enabling high precision.

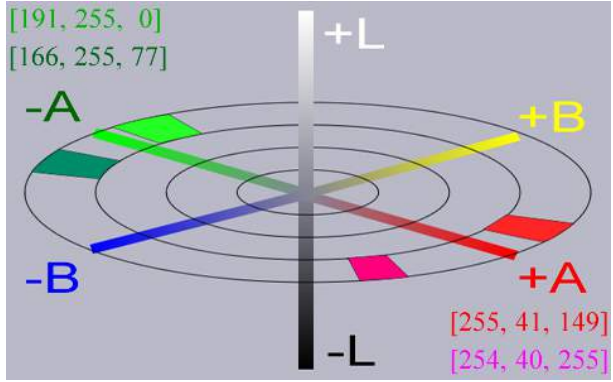

Figure 5: The LAB colorspace. ChromaTag uses red to green borders because they have a large image gradient in the A space. ChromaTag uses two shades of red and two shades of green as representations of 0 and 1 to embed a code in the tag. These different shades of red and green are only differentiable in the B channel, so detection is done in the A channel and decoding is done in the B channel. RGB values for the different red and green colors are shown.

We found LAB to be more robust to color printer and lighting variations than other color spaces such as YUV. We use hashing to speed computation of LAB values.

### 3.2. Precise Localization

Precise localization is required to decode the tag and recover camera pose. Chrominance has lower effective resolution than luminance due to the Bayer pattern filters used in common cameras. ChromaTag is designed with outer black and white concentric rectangles to provide high contrast and high resolution borders for precise corner localization.

## 4. ChromaTag Detection

Algorithm 1 outlines ChromaTag detection. Substantial effort was required to make each step of the detection computationally efficient (700+ FPS!) and robust to variable lighting and perspective.

**FindADiff:** The first step is to find potential tag locations. Assuming that tag cells are at least $N/2$ pixels wide, we search over pixels on a grid of every $N^{th}$ row and column. If a sampled pixel at location $(i, j)$ is in an already-detected area (*InPreviousDetectionArea(i,j,Dets)*), $j$ is moved to the next grid location outside the tag detection (*MoveJToEndOfDetection(i,j)*). Otherwise, the pixel is converted to the LAB space (*ConvertToLAB( Im(i,j) )*), and the A channel intensity is compared to the previous grid location's A channel intensity (*A - OldA > ADiffThresh*). If

---

**Algorithm 1:** ChromaTag Detection

Im = Input RGB image
N = Stepsize in pixels (4 in our experiments)
ADiffThresh = Threshold for initial detection (25 in our experiments)
Dets = Struct to hold detections
TmpDet = Holds detections as they are built

**for** *i=0; i < Im.Rows(); i+=N* **do**
    OldA = ConvertToLAB( Im(i,j) )
    **for** *j=0; j < Im.Cols(); j+=N* **do**
        **if** *InPreviousDetectionArea(i,j,Dets)* **then**
            j = MoveJToEndOfDetection(i,j)
            OldA = ConvertToLAB( Im(i,j) )
            continue

        [L,A,B] = ConvertToLAB( Im(i,j) )
        **if** *A - OldA > ADiffThresh* **then**
            **if** *InitialScan(Im,i,j,A,TmpDet)* **then**
                **if** *BuildPolygon(Im,TmpDet)* **then**
                    **if** *PolyToQuad(Im,TmpDet)* **then**
                          **if** *Decode(Im,TmpDet)* **then**
                              Dets.Add(TmpDet)

    OldA = A

---

the difference is greater than *ADiffThresh*, detection commences and the A value is set as *ReferenceRed*; otherwise, the loop continues to the next sampled pixel. In our experiments, *N=4* and *ADiffThresh=25*.

**InitialScan( Im,i,j,A,TmpDet ):** The next step is to reject red locations that are not tags so that no further processing is done at that location. From the grid location *(i,j)*, we scan left, right, up and down as shown in Figure 6a. Each scan continues until $M$ successive pixels have A channel differences greater than $BorderThresh$ when compared to *ReferenceRed*. If successive pixels are found, scanning has entered the green region and the border *(u,v)* location and pixel value (*ReferenceGreen*) are remembered. If successive pixels are not found, we return *false* and the grid location is abandoned. In our experiments, $M = 3$ and $BorderThresh = 5$.

If all four scans (up, down, left, right) find the green region, we average the *(u,v)* locations on the red-green border to estimate the center of the tag and repeat the scan. If any scan fails, we return *false* and the grid location is abandoned. When the center location converges, we continue the scans through the green region to find the green-black and black-white borders. These scans compare against a set *ReferenceGreen* or *ReferenceBlack* respectively and use the same $M$ and $BorderThresh$ parameters. Center convergence and scans are depicted in Figure 6b. If all scans are successful, the *(u,v)* locations for each border are used to build three initial polygons as shown in Figure 6c.

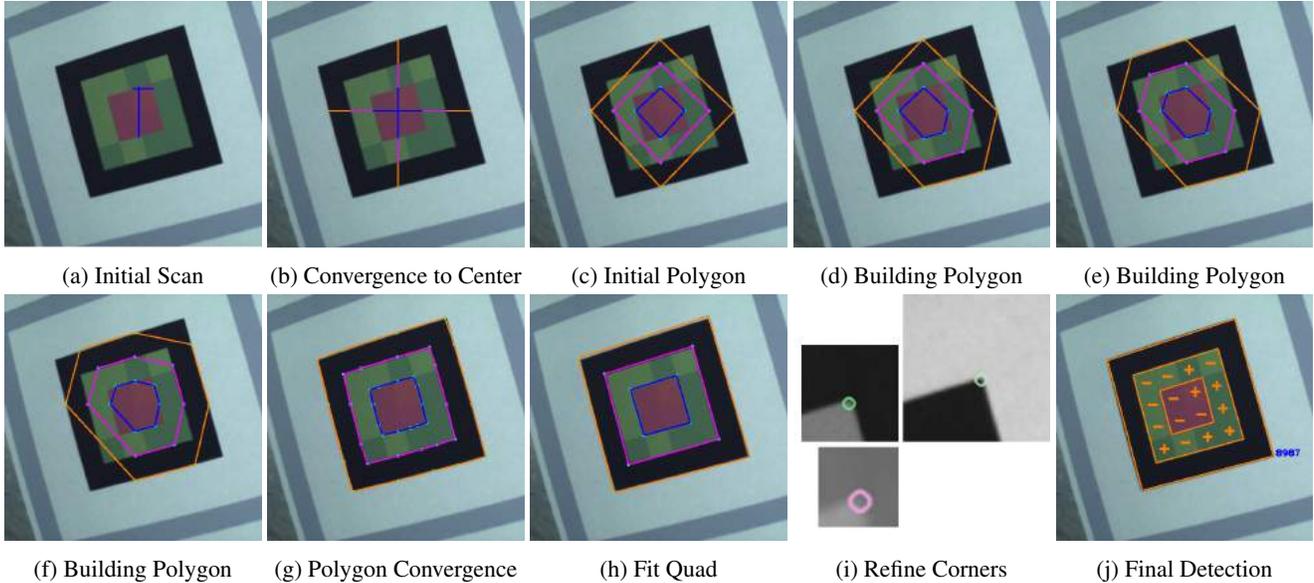

(a) Initial Scan  (b) Convergence to Center  (c) Initial Polygon  (d) Building Polygon  (e) Building Polygon

(f) Building Polygon  (g) Polygon Convergence  (h) Fit Quad  (i) Refine Corners  (j) Final Detection

Figure 6: Scans up, down, left, and right are done to find the red-green border (Figure 6a). If the red-green border is found, the center point is adjusted and scans are repeated until the center point converges. Scans then continue to find all borders (Figure 6b). An initial polygon is built from the found border points (Figure 6c). Additional scans add new points to the polygon (Figures 6d, 6e, and 6f). After enough scans, the polygon converges to the tag quadrilateral (Figure 6g). Polygon edges are clustered into four edges of a quadrilateral (Figure 6h). Patches around each quadrilateral corner are searched for a precise corner location (Figure 6i). A homography is fit and a grid of pixel locations are sampled to decode the tag (Figure 6j).

**BuildPolygon( Im,TmpDet ):** The next step is to expand the initial polygon to match the tag borders. Because squares project to quadrilaterals in the image (ignoring lens distortion), the polygon must remain convex. This limits the maximum potential area that can be added to the polygon with the addition of a new point. Figure 7 shows the maximum area associated with each edge. We build the polygon by greedily scanning in the direction of maximum potential area. The scanning procedure is the same as that described for *InitialScan*. Figures 6d, 6e, 6f demonstrate how iterative scans add points to the polygon along the tag border. Convergence is reached when the ratio of potential areas and polygon area is greater than $ConvThresh$; resulting in a polygon roughly outlining the tag border (Figure 6g). If a scan fails, *false* is returned and the grid location is abandoned. In our experiments, $ConvThresh = 0.98$.

**PolyToQuad( Im,TmpDet ):** Next, a quadrilateral is fit to the polygon. We cluster the angles of the edges weighted by edge length using K-Means ($K = 4$). The cluster centers and outer-most point of each cluster defines four lines and their intersections form the four corners of the quadrilateral (Figure 6h). A patch around each corner is searched using

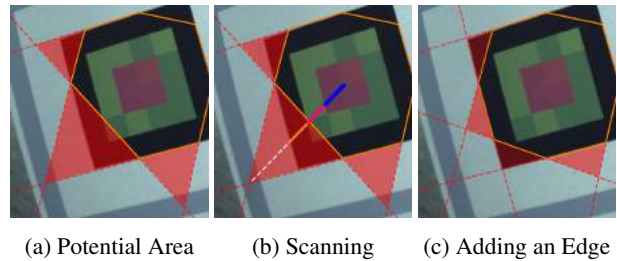

(a) Potential Area  (b) Scanning  (c) Adding an Edge

Figure 7: The potential areas (shown in red) are constructed from neighboring edges extending to an intersection point. The intersection point defines the apex of the triangle and the maximum area that can be added to the current polygon without violating convexity. The next scan moves towards the apex of the largest potential area and finds each border along the way. The new points are added to each border (only outer border is shown).

GoodFeaturesToTrack [32], and the highest scoring point is saved as the new quadrilateral corner. Patch size is scaled by the size of the quadrilateral. Figure 6i shows an example patch and corner for each border.

**Decode( Im,TmpDet ):** A homography matrix is estimated from the black-white border corners. A grid is fit to the black-white border using the homography and the pixel at each grid location is converted to the B channel. The red

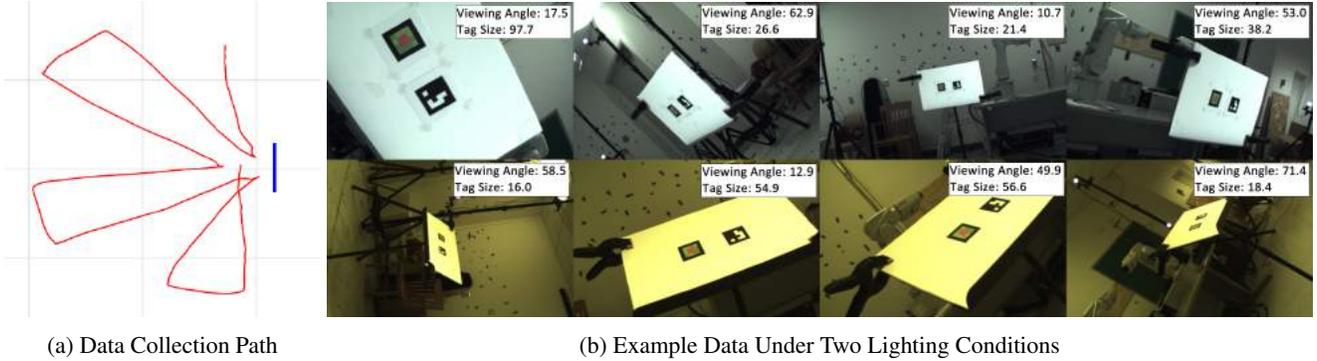

(a) Data Collection Path  (b) Example Data Under Two Lighting Conditions

Figure 8: The tag (blue) and the camera trajectory (red) are shown. Example images annotated with tag size (square root of area) and viewing angle are shown in Figure 8b (top row: white balance (WB); bottom row: no white balance (NWB)).

and green pixels are each clustered separately where the decision boundary is the midpoint of the max and min. The clusters represent the 1 and 0 values that define the tag signature. Figure 6j shows the grid of samples that were decoded in order to finalize detection of the tag. For clustering to work, both shades of red and and both shades of green must be represented in the tag; we remove from established tag libraries any codes that do not satisfy this requirement. The decoded signature is identified as a match using a pre-computed hash-table containing all the signatures as was done for AprilTag [33]. If a match is not found, *false* is returned and the grid location is abandoned.

## 5. Results and Discussion

We collect six datasets for comparison of ChromaTag against AprilTag [33], CCTag [8], and RuneTag [4]. Colored images are captured at 30 fps with a resolution of 752 x 480 using a Matrix Vision mvBluefox-200wc camera [3]. The camera is attached to a servo motor that continuously rotates the camera inplane between 0 and 180 degrees. For each dataset, ChromaTag and one of the comparison tags is placed side-by-side on a flat surface. Data is captured as the camera is moved around the scene. Both tags remain in the image frame during the entirety of the captured sequence. A similar trajectory is traversed three times as the pitch of the tag surface is adjusted between 0 degrees (vertical), 30 degrees, and 60 degrees. A motion capture system is used to capture the pose of the camera during the data collection. This collection is repeated twice for two different lighting conditions: white balance (WB) and no white balance (NWB). Figure 8 depicts the path that is walked and some sample images. The same ID from the 16H5 family was used for both ChromaTag and AprilTag. Tag locations are hand annotated in each image. Additional information about the data can be found in the supplemental material.

The implementations provided by the authors of April-Tag, CCTag, and RuneTag are used for all experiments. All processing uses a 3.5 GHz Intel i7 Ivy Bridge processor.

|          | Average Frames Per Second |||
|----------|-------|---------------|--------------|
|          | Total | > 0 Detections | 0 Detections |
| ChromaTag | 926.4 | 709.2 | 2616.1 |
| AprilTag  | 56.1  | 56.3  | 49.0   |
| CCTag     | 10.0  | 6.5   | 18.5   |
| RuneTag   | 41.9  | 2.4   | 71.3   |

Table 1: ChromaTag has a faster average frames per second for all frames, frames with at least one detection, and frames with 0 detections.

| Detection Step | Average Time Spent on Each Step |
|----------------|--------------------------------|
| FindADiff      | 0.52 ms |
| InitialScan    | 0.03 ms |
| BuildPolygon   | 0.08 ms |
| PolyToQuad     | 0.74 ms |
| Decode         | 0.04 ms |

Table 2: For frames with at least one detection, *PolyToQuad* and *FindADiff* dominate computation.

### 5.1. Detection Speed

**ChromaTag's detection algorithm is faster than other state of the art tags.** Table 1 provides the computation time for each fiducial marker. Table 3 provides the number of frames timed for each marker. The average frames per second (FPS) was calculated by dividing 1 by the mean of the computation time. From Table 1, we see that ChromaTag achieves an average FPS of 926.4, which is 16x, 92x, and 22x faster than AprilTag, CCTag, and RuneTag respectively.

The frames with detection often require more computation than those without. Thus, we provide the computation times for correct detections (true positives). Table 1 shows that ChromaTag has an Average FPS of 709.2 for true pos-itive detections, which is 12x, 109x, and 295x faster than AprilTag, CCTag, and RuneTag respectively.

|          |     | Chroma | April | Chroma | CCTag |
|----------|-----|--------|-------|--------|-------|
| Frames   | WB  | 10266  |       | 11238  |       |
|          | NWB | 10303  |       | 10891  |       |
| Precision| WB  | 96.9   | 46.0  | 96.3   | 100.0 |
|          | NWB | 95.7   | 42.9  | 95.7   | 99.9  |
| Recall   | WB  | 64.0   | 96.4  | 64.5   | 45.7  |
|          | NWB | 67.9   | 98.2  | 66.1   | 46.3  |

Table 3: Each dataset is a pairwise comparison between ChromaTag and another tag with white balance (WB) and no white balance (NWB). Precision and Recall are calculated for each dataset. ChromaTag achieves high precision and better recall than CCTag. AprilTag is successfully detected in almost every frame (high recall); however, many false positives are also detected (low precision).

| Detection Step | Percent of Frames(%) |
|---|---|
| FindADiff    | 2.0  |
| InitialScan  | 2.8  |
| BuildPolygon | 25.8 |
| PolyToQuad   | 1.6  |
| Decode       | 1.1  |

Table 4: For each step, the % of frames that failed is shown. *BuildPolygon* is most often the step where failure occurs.

The frames without detections should require very little time because time spent on tagless images or failed detections is wasteful. We calculate computation time for frames without a detection (false negatives because all data contains the tags). ChromaTag has an average FPS of 2616.1, which is 37x faster than the next fastest tag (RuneTag).

Table 2 shows how much time each step of ChromaTag detection uses. Only frames where successful detection occurred are used for this breakdown. *PolyToQuad* (0.74 ms) and *FindADiff* (0.52 ms) are the most costly. *PolyToQuad* is costly because of K-Means clustering and corner localization, and *FindADiff* is costly because of scanning the image and converting pixels to the LAB color space.

## 5.2. Detection Accuracy

Table 3 summarizes the detection results for ChromaTag compared to AprilTag and CCTag. We define true positives (TP) as when the tag was correctly detected in the image. This means locating the tag and correctly identifying the ID. Correctly identifying the tag is determined by having at least 50 percent intersection over union between the detection and the ground truth (though detections by all tags far exceeds this threshold). We define false positives (FP) as detections returned by the detection algorithms that do not identify the location and ID correctly. We define false negatives (FN) as any marker that was not identified correctly. Precision is $\frac{TP}{TP+FP}$ and recall is $\frac{TP}{TP+FN}$.

RuneTag is omitted from Table 3 because it was not detected in our dataset. The maximum size of any tag in the images is about 126x126 pixels. With additional data, we found that RuneTag requires larger tag sizes for detection. See the supplementary material for more information.

Table 4 breaks down how often each step causes failed detection. Most detection failures occur during the *Build Polygon* step (25.8%). Profiling failures is useful to emphasize areas for future improvements.

**ChromaTag's initial false positive rejection improves precision.** Table 3 shows that ChromaTag ( 96%) has a higher precision than AprilTag ( 44%). This is interesting because both tags are using the 16H5 family. ChromaTag is successfully rejecting many initial false positives that April-Tag identifies as tags. Note that the 36H11 family causes less false positives [33]; however, less space on the tag is available for the border and inner grid squares (resulting in less recall). Thus, ChromaTag is able to take advantage of the larger grid and border areas of the 16H5 codes without suffering from false positives like AprilTag. ChromaTag and CCTag have similar precisions (greater than 95%).

**The combination of high recall and fast detection makes ChromaTag a good choice for many applications.** Table 3 shows that ChromaTag has lower recall than April-Tag (and higher recall than CCTag). However, figure 9a shows that for tag sizes (square root of tag area in the image) greater than 70 pixels, ChromaTag achieves similarly high recall. To put that in perspective, the resolution of the dataset images is 752x480 pixels, so a 70x70 tag area is less than 2% of the total available area in these images. Thus, for applications where detecting very small tags is not important, the 16x speed gain makes ChromaTag the better option. Compared to CCTag, ChromaTag achieves similar recall for large tags and higher recall as tag size decreases.

**ChromaTag localizes corners precisely.** Since use of tags for pose estimation depends on accurately localizing the corners, we evaluate accuracy for corner localization compared to hand-labeled corners that are locally refined with a Harris corner detector [20]. We found that ChromaTag localized 94.4% of the corners within 3 pixels of the ground truth corners and AprilTag localized 89.1% of the corners within 3 pixels (94.2% within 4 pixels). This demonstrates that our white-black border design and detection enables precise corner localization on par with that of AprilTag. On visual inspection, errors of within 5 pixels are attributable to reasonable variation.

**ChromaTag is robust to color variation.** Table 3 shows that ChromaTag achieves similar recall for both the WB and NWB datasets despite the colors being significantly different. Specifically, ChromaTag recall is 64.0% and 67.9%

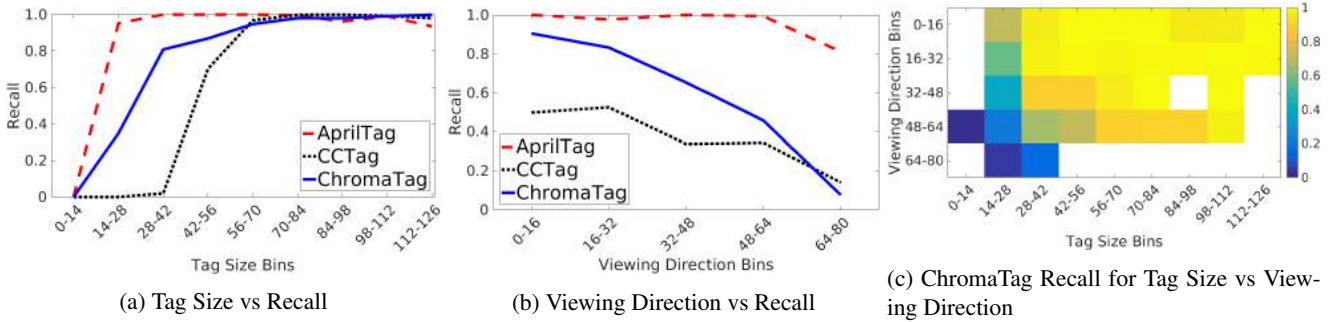

(a) Tag Size vs Recall　　(b) Viewing Direction vs Recall　　(c) ChromaTag Recall for Tag Size vs Viewing Direction

Figure 9: Recall is high when tag size (square root of tag area in image) is large and decreases as tag size decreases (Figure 9a). Recall is high when the camera is frontally facing the tag (low values) and decreases as the angle increases (Figure 9b). Viewing angle is the angle between the normal vector of the tag plane and the camera viewing direction (tag plane center to camera center). ChromaTag achieves similar recall to AprilTag for tags larger than 70 pixels. Figure 9c shows the recall for ChromaTag as it depends on both viewing direction and tag size. Yellow means a high recall and blue means a low recall. We see from this plot that both viewing direction and tag size affect ChromaTag recall, though tag size has a larger direct effect.

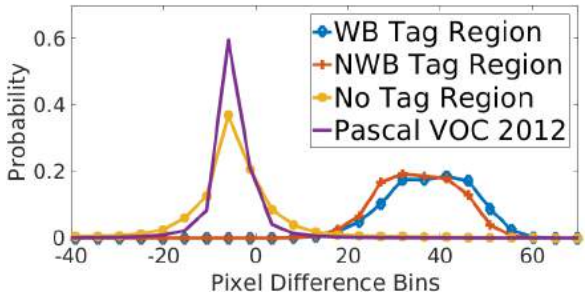

Figure 10: WB Tag Region (blue) and NWB Tag Region (orange) are the regions of the WB and NWB data respectively that include the tag. No Tag Region (yellow) is the tagless parts of the WB and NWB data. Each histogram bins A channel pixel differences from horizontally scanning the image (same approach as our detection algorithm). For WB and NWB Tag Region, only the max pixel difference is binned. The result is two modes highlighting how the pixel difference on the tag is easily differentiable from the pixel differences of natural images. The colors of WB and NWB are drastically different, yet still differentiable from natural images, showing that red and green pixel differences in the A channel are robust to color variation.

for WB and NWB on the AprilTag dataset and $64.5\%$ and $67.9\%$ for WB and NWB on the CCTag dataset. The comparisons for ChromaTag are similarly close for precision between WB and NWB for each dataset. These results provide evidence that ChromaTag is robust to color variation.

ChromaTag is robust to color variation because initial detection and finding borders rely on LAB pixel differences, which the tag design ensures are consistently large. Figure 10 shows histograms of pixel differences in the A channel for the tag region of the WB data (blue), the tag region of the NWB data (orange), the tagless regions of WB and NWB (yellow), and the Pascal VOC 2012 dataset [14] (purple). Each histogram is created by binning A channel pixel differences from horizontally scanning the image with $N = 4$ subsampling (same approach as our detection algorithm). For the tag regions, only the max difference is counted since only one difference must be above threshold for detection. Despite large variation in color between WB and NWB, the A channel difference for tag regions is clearly differentiable from A channel differences in natural images, which shows why ChromaTag is robust to lighting and color variation and can reject false positives quickly.

## 6. Conclusions

We present a new square fiducial marker and detection algorithm that uses concentric inner red and green rings to eliminate false positives quickly and outer black and white rings for precise localization. We demonstrate on thousands of real images that ChromaTag achieves detection speeds significantly faster than the current state of the art while maintaining similar detection accuracy. We also show that ChromaTag detection fails at far distances and steep viewing angles and recommend AprilTag as a better option for applications that require detection in these conditions. Lastly, we provide evidence that ChromaTag detection is robust to color variation, and break down which steps of the detection algorithm take the most time and fail most often.

## Acknowledgment

This work is supported by NSF Grant CMMI-1446765 and the DoD National Defense Science and Engineering Graduate Fellowship (NDSEG). Thanks to Austin Walters, Jae Yong Lee, and Matt Romano for their help with tag design and data collection.